\documentclass[10pt,twocolumn,letterpaper]{article}

\usepackage{cvpr}
\usepackage{times}
\usepackage{epsfig}
\usepackage{graphicx}
\usepackage{amsmath}
\usepackage{amssymb}
\usepackage{comment}
\usepackage{subcaption}

\usepackage[breaklinks=true,bookmarks=false]{hyperref}

\cvprfinalcopy 

\def\cvprPaperID{****} 

\begin{document}

\title{Playing Go without Game Tree Search Using Convolutional Neural Networks}

\author{Jeffrey Barratt\\
Stanford University\\
353 Serra Mall, Stanford, CA 94305\\
{\tt\small jbarratt@cs.stanford.edu}
\and
Chuanbo Pan\\
Stanford University\\
353 Serra Mall, Stanford, CA 94305\\
{\tt\small chuanbo@cs.stanford.edu}
}

\maketitle

\begin{abstract}
The game of Go has a long history in East Asian countries, but the field of Computer Go has yet to catch up to humans until the past couple of years. While the rules of Go are simple, the strategy and combinatorics of the game are immensely complex. Even within the past couple of years, new programs that rely on neural networks to evaluate board positions still explore many orders of magnitude more board positions per second than a professional can. We attempt to mimic human intuition in the game by creating a convolutional neural policy network which, without any sort of tree search, should play the game at or above the level of most humans. We introduce three structures and training methods that aim to create a strong Go player: non-rectangular convolutions, which will better learn the shapes on the board, supervised learning, training on a data set of 53,000 professional games, and reinforcement learning, training on games played between different versions of the network. Our network has already surpassed the skill level of intermediate amateurs simply using supervised learning. Further training and implementation of non-rectangular convolutions and reinforcement learning will likely increase this skill level much further.
\end{abstract}

\section{Introduction}
In all perfect information, discrete games, there exists a policy which takes in the current game state and returns the optimal move for that state for the current player, which will result in the highest utility under perfect play. Although neural networks can theoretically approximate any function, including this policy, doing so in practice is essentially impossible. The ancient board game “Go” is one such perfect information game which has been studied for years in computer science and has been one of the hardest turn-based perfect information games to get good performance in.

\begin{figure}[h]
\centering
\captionsetup{justification=centering}
\begin{subfigure}{.45\linewidth}
  \centering
  \includegraphics[width=.6\linewidth]{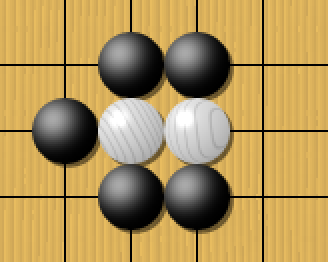}
  \small \caption{The two white stones can be captured, because there is only one free adjacent spot.}
  \label{fig:sub1}
\end{subfigure}
\ \ 
\begin{subfigure}{.45\linewidth}
  \centering
  \includegraphics[width=.6\linewidth]{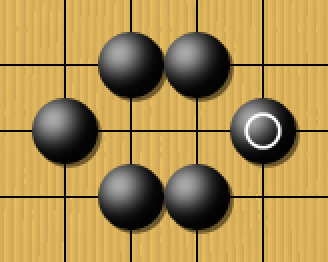}
  \small \caption{The two stones were captured by playing at the circled point.}
  \label{fig:sub2}
\end{subfigure}
\caption{Capturing stones.}
\label{fig:eggame}
\end{figure}

A game of Go consists of two players alternating placing stones on a 19 by 19 board, with the player with black stones starting. Any stones that are directly adjacent to each other are part of the same group. If any group is completely surrounded by opponent stones, it is taken off of the board, as shown in Figure \ref{fig:eggame}. The goal of the game is to surround as much territory as possible.

While the rules of go are simple, mastering the game is not. Almost all commercial Go playing programs rely heavily on some form of tree search to explore outcomes of certain moves in a game, most commonly Monte Carlo Tree Search (MCTS) \cite{gelly2012grand}. However, these programs don't seem to fully understand the game, instead relying on brute force search to make good moves.

In the past two years, many companies have tried to create their own Go playing programs, such as DeepZen, FineArt, and, most notably, DeepMind's AlphaGo. Each of these programs uses a combination of convolutional neural networks and Monte Carlo Tree Search to obtain a level of play at or above top professionals \cite{SilverHuangEtAl16nature}. However, like  commercial go players, these programs are exploring millions of states per second across many computers and countless GPUs, far surpassing the number of positions calculated out by the human player they are playing against.

\section{Related Work}

Computer Go, the creation of Go-playing agents for computers, has existed as early as 1968 \cite{firstcompgo}. As mentioned previously, before convolutional neural networks became popular, MCTS was the most powerful method to play Go. These techniques generally required a lot of enhancements and optimizations. For example, MCTS Solver to detect forced as described by \cite{Winands:2008:MTS:1428170.1428173} allowed Go bots to quickly solve games. Rapid Action Value Estimation used in \cite{Rimmel2011} allowed bots to identify duplicate game states, thereby significantly reducing the number of computations necessary. The problem with these techniques was that there were too many states to evaluate. Therefore, programs such as introduced in \cite{enzenberger2010fuego} and \cite{lee2009computational} were only truly competitive on $9 \times 9$ boards while only capable of achieving moderate success on $19 \times 19$ boards.

The field incrementally grew steadily in a similar fashion until 2015, when AlphaGo first beat Fan Hui, a professional player, making a huge skill jump that hadn't yet been seen \cite{SilverHuangEtAl16nature} in Computer Go. This led to a huge boom in convolutional neural network-based architectures such as DeepZenGo, FineArt, and of course AlphaGo \cite{dzfaag}. The inspiration for using convolutional neural networks comes from the fact that the state of a Go board can be treated as an image. While they rely on a policy network much like this paper describes, their strength comes from exploration of millions of board positions to gain an advantage over their human opponents.

Previous work has been done on using only convolutional neural network to play Go. They offered boosts over traditional MCTS but were not able to achieve the same level of play as AlphaGo. In 2008, \cite{Sutskever2008} created a Convolutional Neural Network to play Go using an ensemble of networks. They were only able to achieve a then state-of-the-art 36.9\% accuracy with a relatively small number of parameters ($10^4$).\cite{shitversion}, in 2014, improved this to 41 and 44\% validation accuracy on different data sets. Concurrently, \cite{DBLP:journals/corr/MaddisonHSS14} devised a deeper and larger network that achieved 55\% accuracy. The first two papers used supervised learning, tied weights, and a relatively shallow network, which resulted in accuracies much below what we and the creators of AlphaGo and \cite{DBLP:journals/corr/MaddisonHSS14} have observed. \cite{DBLP:journals/corr/MaddisonHSS14} had a model around the size of AlphaGo's policy network (13 convolution layers), and had comparable results.

More recent approaches such as \cite{DBLP:journals/corr/TianZ15} have introduced fancier convolutional neural networks that rely on long term predictions for extended play. \cite{DBLP:journals/corr/TianZ15} achieved a slightly better accuracy of around 56\%-57\% and placed 3rd overall at the KGS Computer Go Competition.

It's important to remember that deep convolutional neural networks are not just used to play Go. Many games such as Chess, Stratego, Hexagon and, more obviously, Atari games can be treated as images with labels being where to move next \cite{lewisplaying} \cite{smithlearning} \cite{oshripredicting} \cite{DBLP:journals/corr/MnihKSGAWR13}. This shows the flexibility of deep convolutional neural networks as a tool to model many hard to play (and hard to understand) games.

\section{Problem Statement}
The goal of this project is to create human-level understanding of the game by creating a player which does no tree search whatsoever; to simply rely on understanding the board position and ``intuition'', rather than brute force calculation. We plan to train a convolutional neural network which tries to approximate the previously discussed optimal policy function and as a result create a strong go player which is able to beat traditional MCTS-based programs.

The rankings in Go scale from 25 kyu (worst rank) to 1kyu (best kyu rank), and from 1 dan to 9 dan (best rank). The 1 dan rank is one rank above the 1 kyu rank. Hopefully, a rank of at least 1 dan can be achieved with our program, tested through websites such as the Online Go Server (OGS), as well as with in-person matches to ballpark its strength. Playing our player against existing go programs with known ranks can also be a good way of determining an approximate rank for our player.


\section{Methods}

\subsection{Cross-Shaped Convolutions}
\begin{figure}[h]
\centering
\includegraphics[scale=0.1]{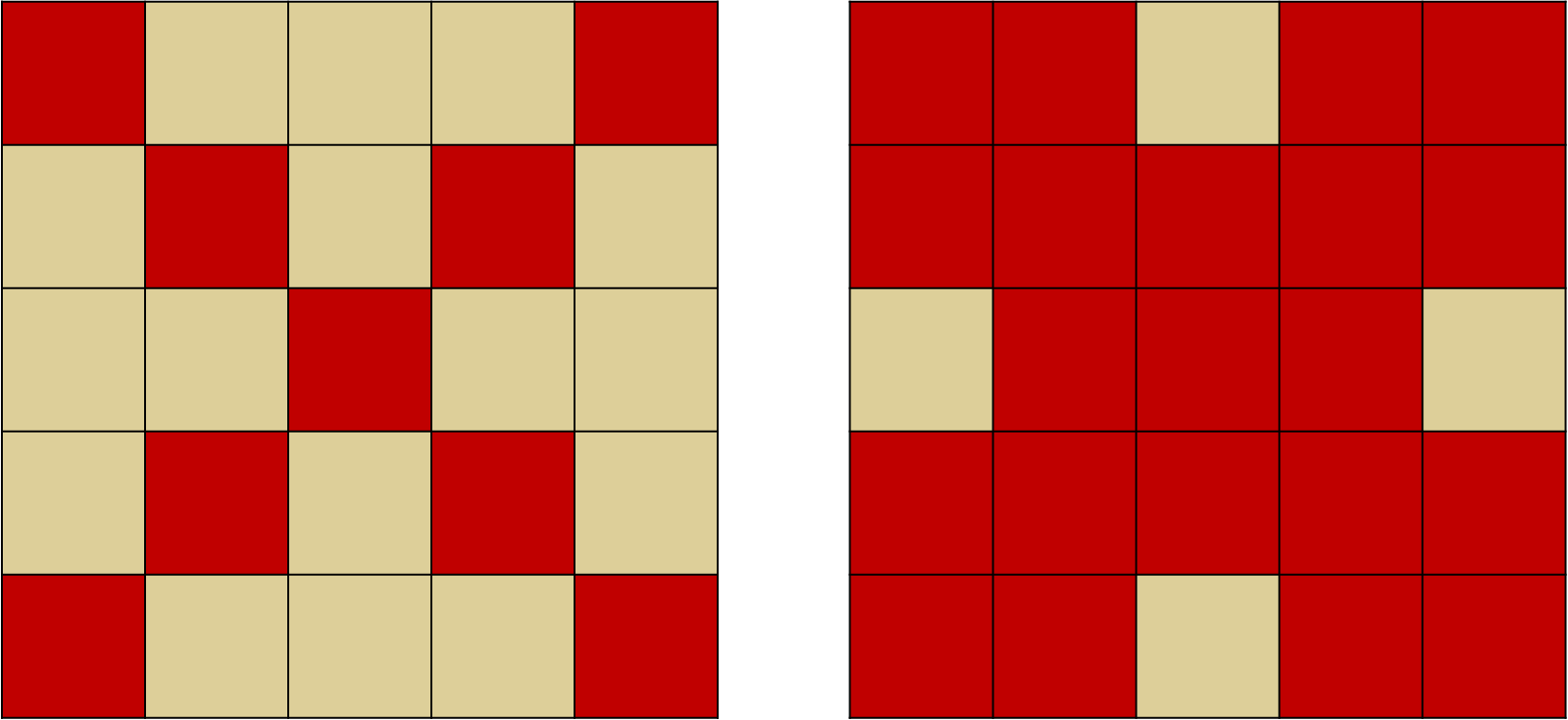}
\caption{$5\times5$ convolutions with cross-widths of 1 and 2 respectively.}
\label{fig:crossconv}
\end{figure}
In addition to rectangular convolution filters, we implemented a novel cross-shaped convolution filter as shown in Figure \ref{fig:crossconv}. 

\begin{figure}[h]
\centering
\captionsetup{justification=centering}
\begin{subfigure}{.45\linewidth}
  \centering
  \includegraphics[width=.6\linewidth]{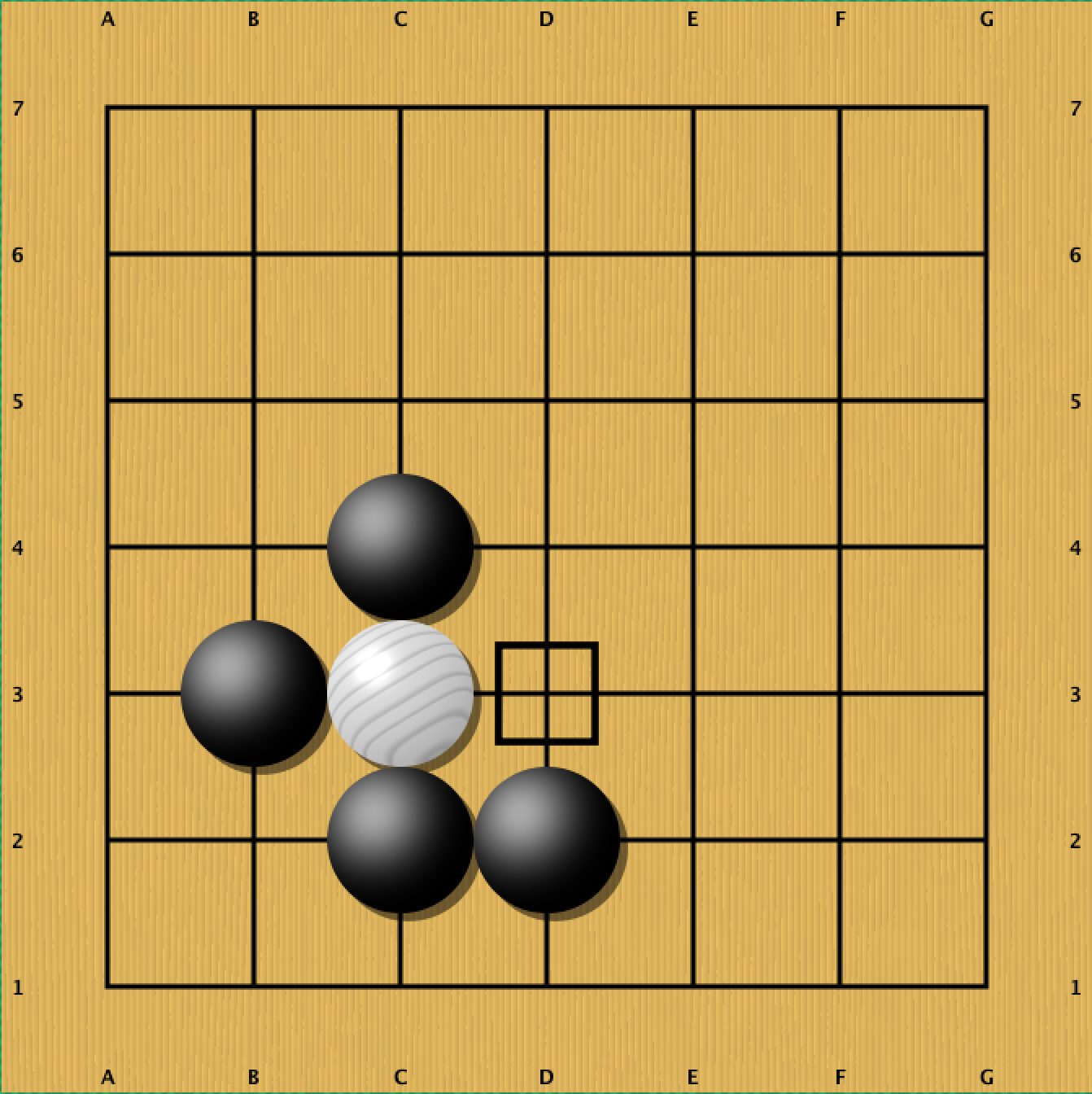}
  \small \caption{If the white player wants to save the stone, she must play on the squared point.}
  \label{fig:sub1}
\end{subfigure}
\ \ 
\begin{subfigure}{.45\linewidth}
  \centering
  \includegraphics[width=.6\linewidth]{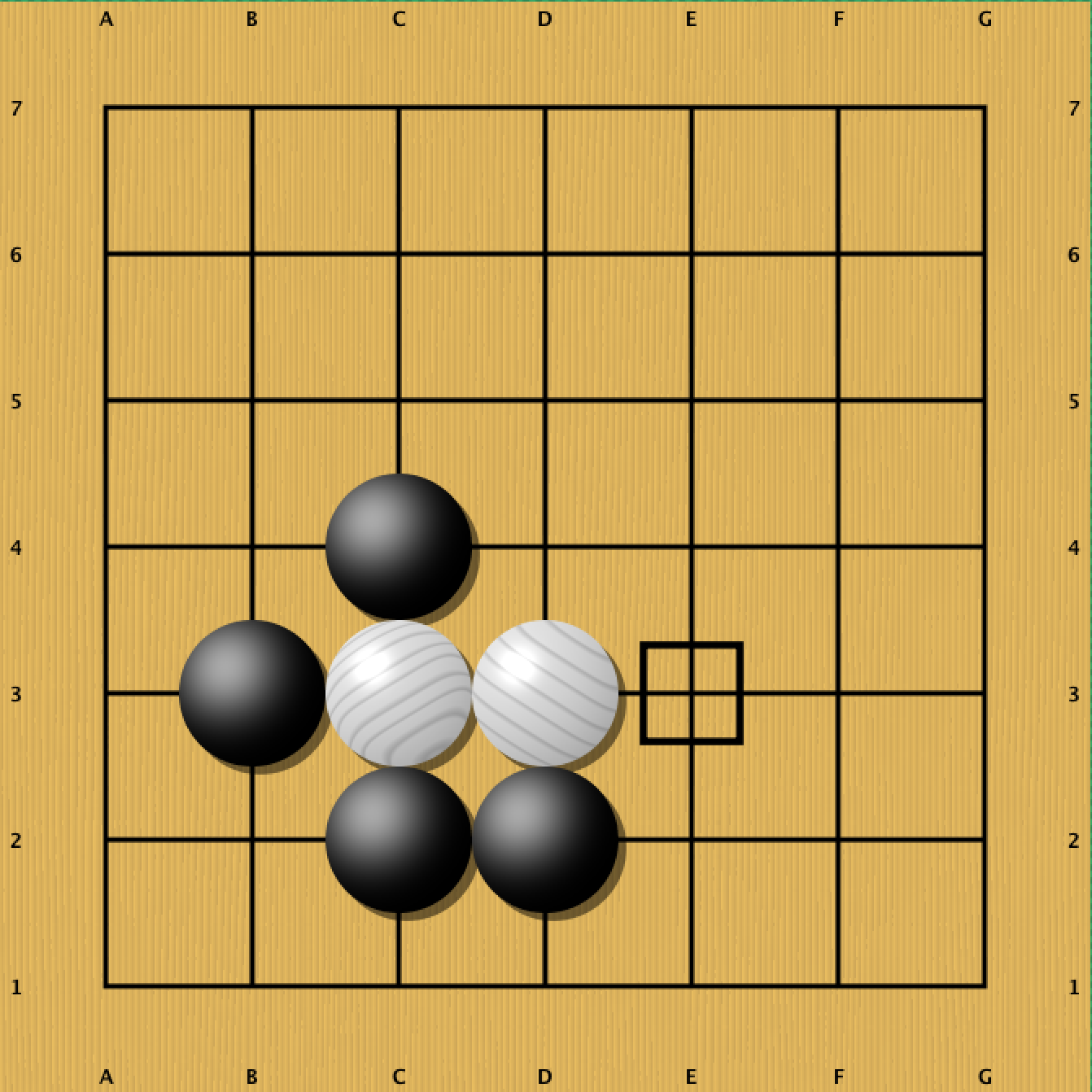}
  \small \caption{The black player can eventually capture by playing on the squared point.}
  \label{fig:sub2}
\end{subfigure}\\
\begin{subfigure}{.45\linewidth}
  \centering
  \includegraphics[width=.6\linewidth]{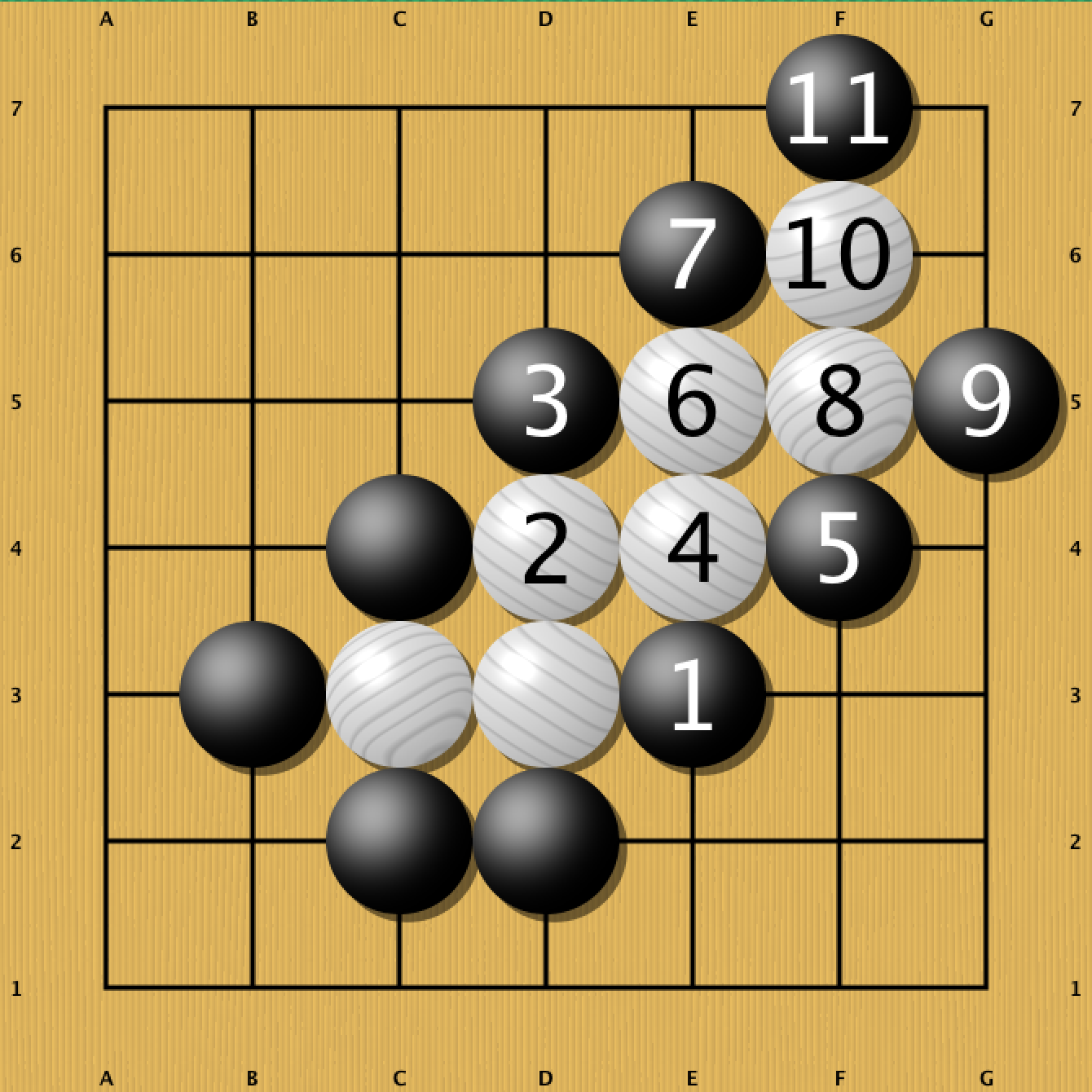}
  \small \caption{Attempts at escape are useless, as seen in moves 1-11 above.}
  \label{fig:sub1}
\end{subfigure}
\ \ 
\begin{subfigure}{.45\linewidth}
  \centering
  \includegraphics[width=.6\linewidth]{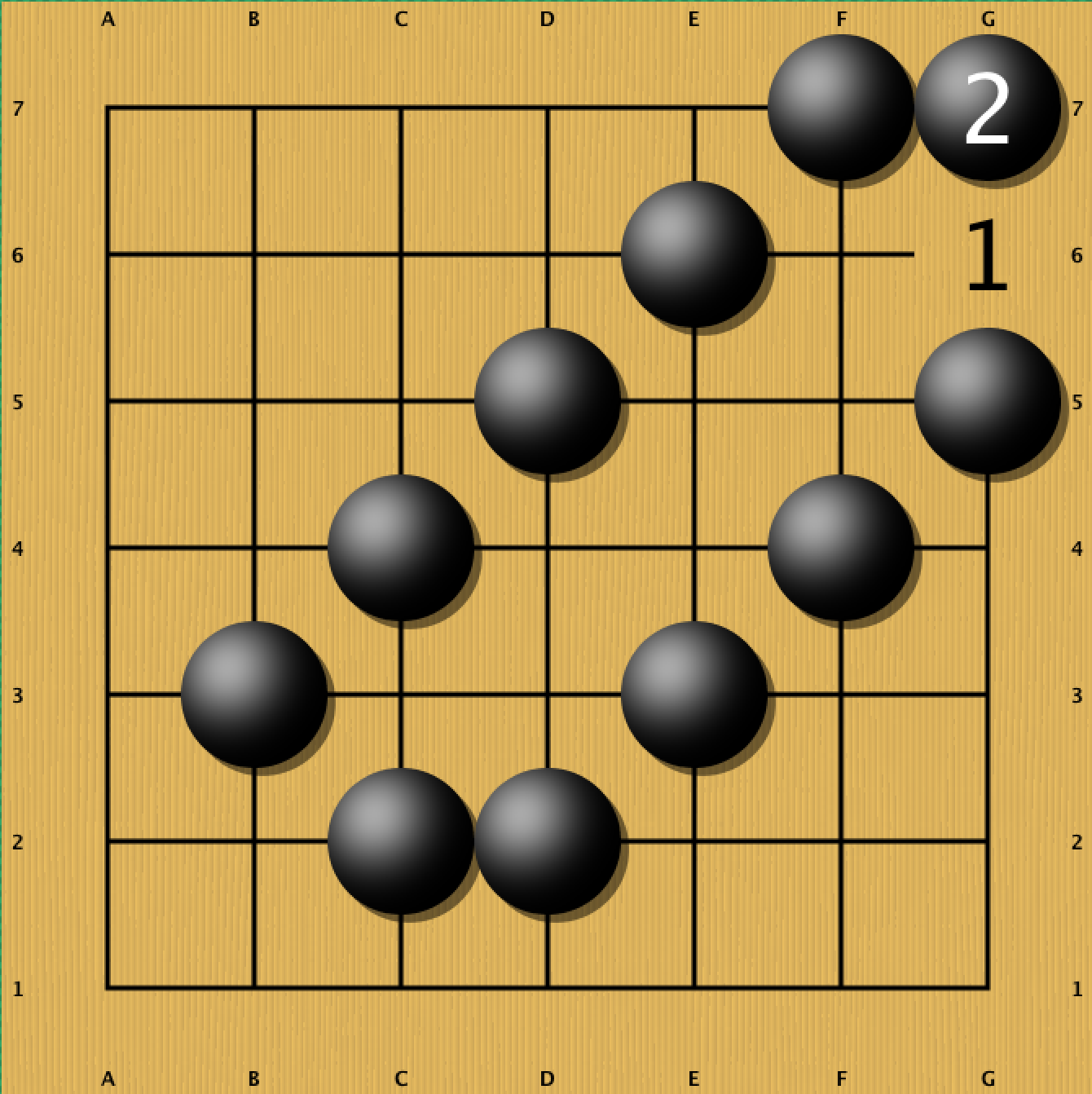}
  \small \caption{The white stones will eventually be captured, if white continues to play.}
  \label{fig:sub2}
\end{subfigure}
\caption{An example ladder on a 7x7 board.}
\label{fig:egladder}
\end{figure}

These convolutions were motivated by patterns in Go called ``Ladders''. Ladders are a capturing concept in Go that only appear in low-amateur level play, with some exceptions \cite{TheLadderGame}, but are considered in all levels of play. They are never played out because they are disastrous for one of the players involved to continue. The ladder shape continues across the board in a diagonal fashion, as seen in Figure \ref{fig:egladder}.

\begin{figure}[h]
\centering
\captionsetup{justification=centering}
\begin{subfigure}{.45\linewidth}
  \centering
  \includegraphics[width=.6\linewidth]{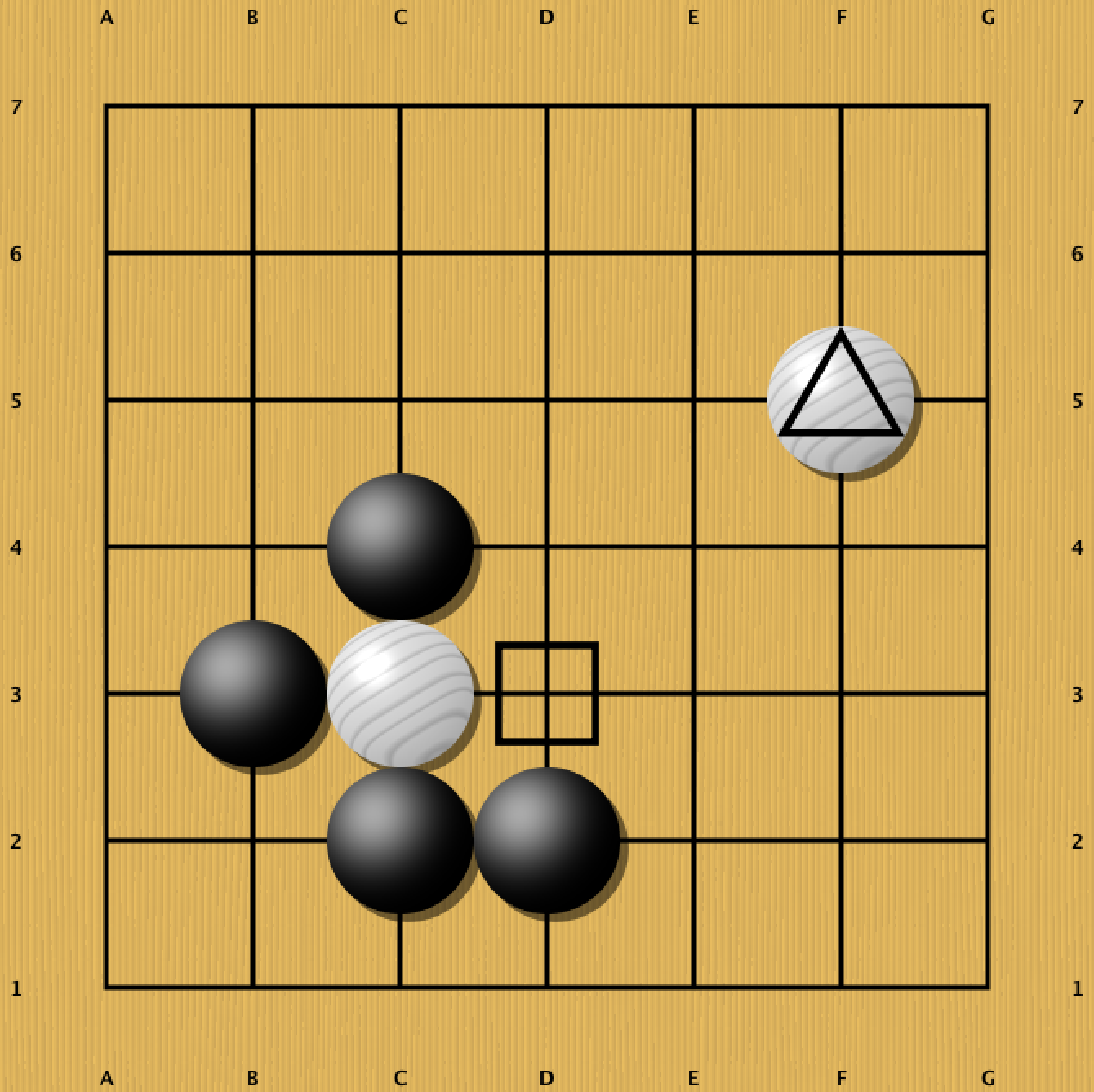}
  \small \caption{The ladder situation is different here, because white has a stone in the way, marked with a triangle.}
  \label{fig:blsub1}
\end{subfigure}
\ \ 
\begin{subfigure}{.45\linewidth}
  \centering
  \includegraphics[width=.6\linewidth]{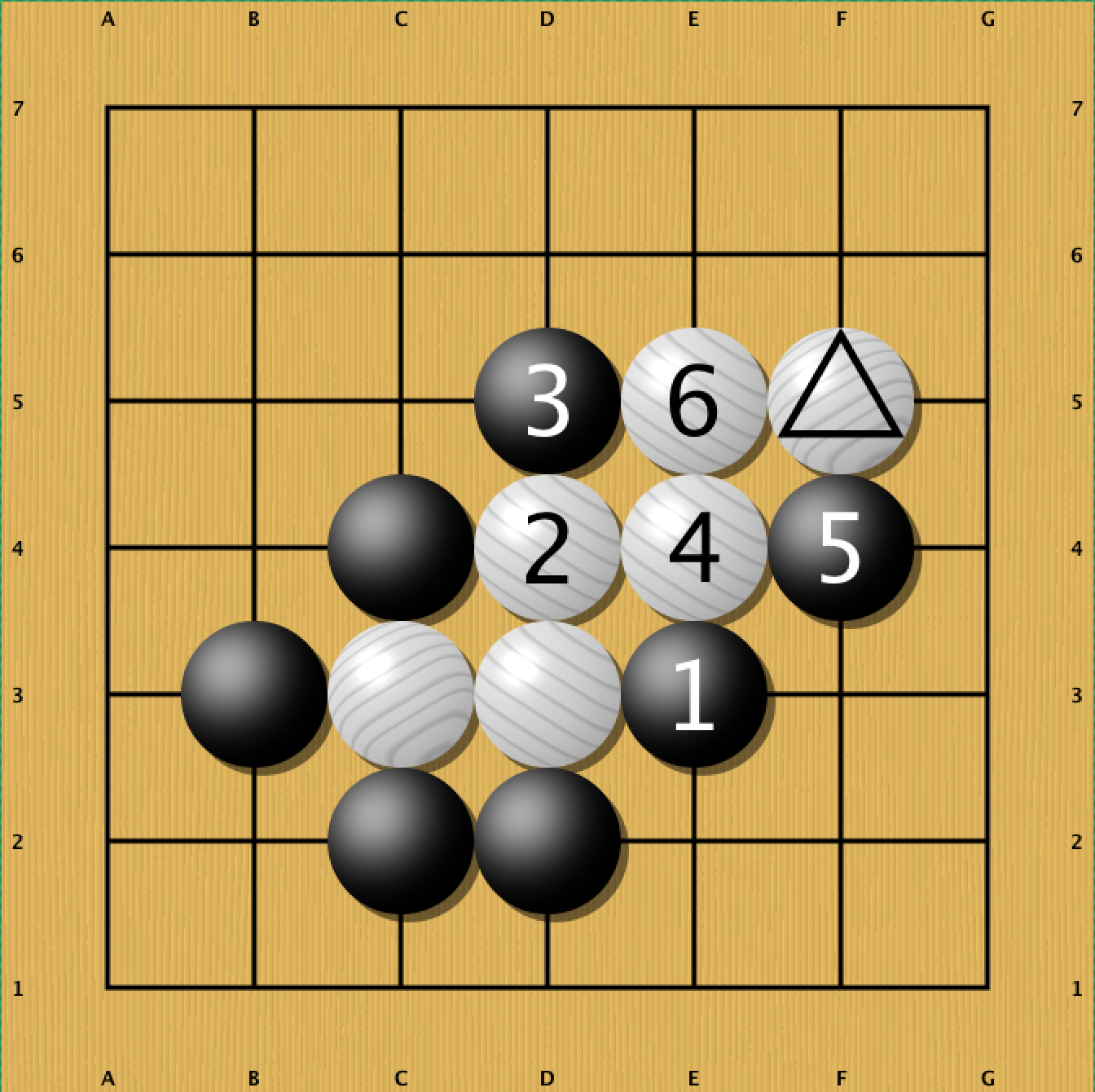}
  \small \caption{The ladder, if played out, results in white's escape. An attempt at capture failed for black.}
  \label{fig:blsub2}
\end{subfigure}
\caption{Capturing stones.}
\label{fig:egbrokenladder}
\end{figure}

However, if a white stone is present, the ladder can be \textit{broken}, allowing white to escape, as seen in Figure \ref{fig:egbrokenladder}. The board position in all diagonal directions must be considered in a game of go, and thus we aimed to solve this problem by developing a diagonal-shaped convolution.

These cross convolutions are built inside traditional $n \times n$ convolutions with the unnecessary weights zeroed out. To further describe cross convolutions, we first define the notion of the cross-width and cross-area. The cross-width, $c$, of a cross convolution is the thickness of the line as it spans from one corner to another. The cross-area is defined as the set $S_{CA}$ that contains all the coordinates covered by a $c \times c$ block moving from one corner to the opposite corner. The cross-area contains the active regions of the convolution. The cross width must fall within the range $1 \leq c \leq \lceil n/2 - 1\rceil$. Anything beyond this range will create a zero or normal convolution respectively.

In practice, it's easier and to zero out values based off of points that are not in $S_{CA}$ because a $c \times c$ block moving from one corner to another tends to create a lot of overlap. We realize this task is made even easier by the fact that the points not in $S_{CA}$ simply form four triangles on each side. Our problem is therefore simplified down to creating a filter that zeros out the coordinates of the four triangles. Once the filter has been created, it is then used throughout the duration of the model to ensure only the relevant weights are utilized.

By creating these cross convolutions, we introduced the cross-width as another hyper-parameter that we can tune. Adjusting these hyper-parameters determines how our player visualizes the board with respect to the ladder structure. Previous non-MCTS based techniques usually represent ladders and other structures as feed-input features (and only these features), but we wish to focus on representing the general state of the board so our player can play effectively from beginning to end \cite{Graepel2001}. However, we might consider analyzing patterns as \cite{Herbrich15} does and inserting another layer in a way that doesn't conflict with our current inputs.

\subsection{Cross Layer}
\begin{figure}[h]
\centering
  \includegraphics[width=.7\linewidth]{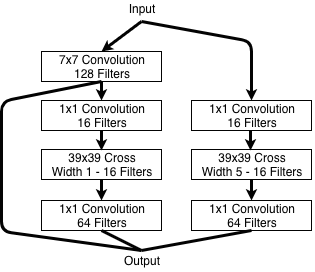}
  \caption{A single cross convolutional layer.}
  \label{fig:model}
\end{figure}
We did not directly incorporate our cross convolution directly in the network as its own individual layer. Instead, inspired by Google's Inception Module as shown in \cite{Szegedy_2015_CVPR}, we decided to incorporate our model in a modular ``cross layer'' as shown in Figure \ref{fig:model}. Cross convolutions are generally very sparse. Therefore using cross convolutions in isolation would more destructive than constructive. As such, the cross layer introduces the cross convolution as a concatenation to our original filters. Therefore, we do not loose any meaningful data. Additionally, concatenating the pre-cross convolution input is very reminiscent of ResNet potentially helped in strengthening gradient flow \cite{DBLP:journals/corr/HeZRS15}.

\begin{figure}[h]
\centering
  \includegraphics[width=.56\linewidth]{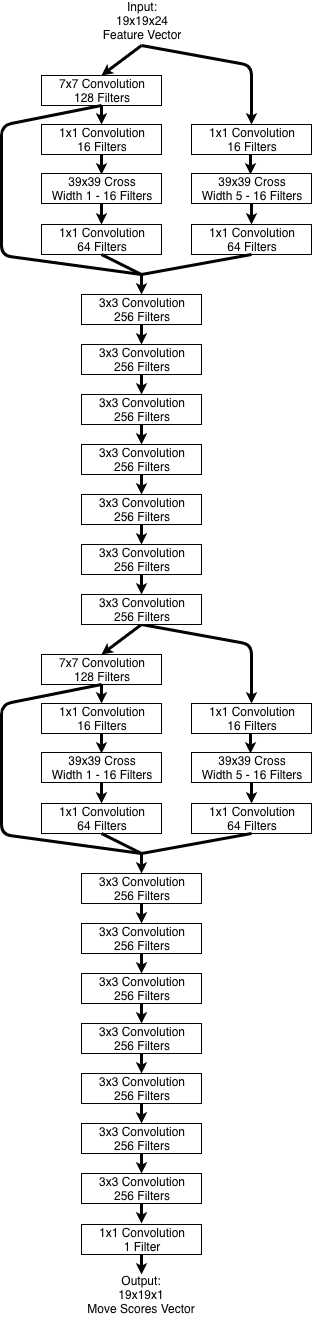}
  \caption{Our model.}
  \label{fig:model}
\end{figure}

\subsection{Model}
Our final model consists of a 23-layer convolutional neural network, using a completely new architecture. Our first layer consists of a $7\times7$ convolutional filter with stride 1, pad 3 and width 128. This feeds directly into the $3\times3$ layers, as well as a 1-width cross convolution, as discussed below. The $1\times1$ convolutions surrounding the $39\times39$ cross convolutions serve to lower the number of parameters in the $39\times39$ layer, a technique similar to the popular SqueezeNet architecture \cite{squeezenet}. The input is also fed directly into a squeeze and 5-width cross convolution layer. All outputs are then concatenated and sent to the next layer. Layers 5-11 are all $3\times3$ convolutional filters with stride 1 and pad 1, each with 256 layers. We add another $7\times7$ convolution and squeeze cross convolutions because there is some evaluation of board positions present in the data once we are through layer 11 that we would like to know for the situation of the ladder. Layers 16-22 are similarly all $3\times3$ convolutional filters with stride 1 and pad 1, each with 256 layers. The final layer is a $1\times1$ convolution with only 1 filter, used to flatten the baord position into a $19\times19$ score vector.

We previously had a fully connected layer to evaluate final board positions. However, we discovered that the a significant portion of training was spent at this layer. Therefore, we instead took the results of the final layer as the scores for each move at the given point on the $19\times19$ board. We select from the list of scores the highest legal move.

\subsection{Training}
We began training our network by method of supervised learning, using the 9 million of the 10 million state-move pairs discussed in section 5: Dataset and Features. 1 million of the 10 million state-move pairs were separated off to be used as a test set. We trained the network using vanilla stochastic gradient descent using a learning rate of $\alpha = 0.001$, and decayed the learning rate by 0.5 every epoch. We ran the training using an NVIDIA 960. Training the network to its current state, over the course of four epochs, took five days.

We were unfortunately out of time by the end of the project to train our network using reinforcement learning. However, future work over this summer will likely be conducted to further strengthen the player in real-game scenarios as opposed to the mimicking strategy currently employed in the supervised learning network.

\subsection{Interfaces and Evaluation}
\begin{figure}[h]
\centering
  \includegraphics[width=\linewidth]{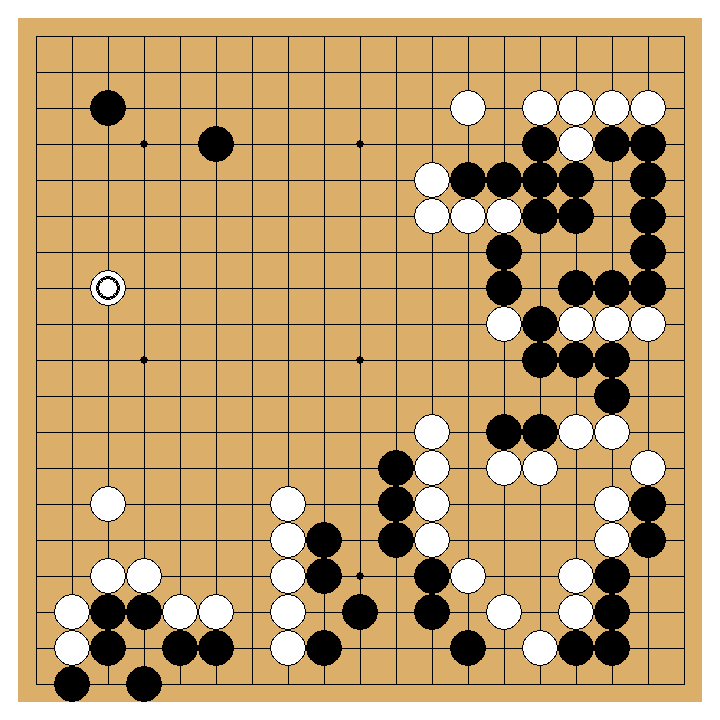}
  \caption{An example position while playing against the network. This interface was developed as part of the project.}
  \label{fig:interface}
\end{figure}
As part of the project, we developed a visual interface to be able to play against the player as well as use it as a tool to evaluate board positions. The interface, shown in Figure \ref{fig:interface}, communicates with the network to both give it the rotational symmetries, as discussed in the next section, and any other board features that are necessary. Once the network finishes its forward pass, the interface iterates through the scores given by the player, removing any illegal moves, and plays the highest-scoring move. We also implemented an interface that shows the top 10 probabilities given to certain moves on the board, as shown in the results section. These implementations allowed us to visualize and understand what our player was ``thinking'' in any given board position, as well as play against it to evaluate its playing strength.

\subsection{Symmetries}
Because all board positions in Go play out the same way when flipped or rotated, our player employs an ensemble of eight passes through the neural network using all eight symmetries of the two-dimensional board. Because we did not employ weights that were locked to each other in a symmetric fashion, as in \cite{shitversion}, this helped our player play in the early game more generally, as some games start in a different corner, but are still equivalent in terms of symmetry.


\section{Dataset and Features}
Our data set consists of 53,000 professional games from as early as the 11\textsuperscript{th} century to 2017 in Smart Game Format (SGF) \cite{GoData}. We implemented parsers using sgfmill \cite{sgfmill}, a python library for simulating sgf games, into 10 million state-move pairs to be used in the training process.

Each state is represented as a $19 \times 19 \times 24$ vector. Each position on the board has 24 one-hot features, as shown in Table 1:
\vspace{2mm}

\noindent \textbf{Table 1}\vspace{1mm}\\
\begin{tabular}{c|c}
\hline
Number of Feature Layers & Feature Layer Contents\\
\hline
1 & Our Stone Present\\
1 & Opponent Stone Present\\
1 & Blank Space\\
1 & Legal Move\\
4 & Liberties After Move Is Played\\
8 & Time Since Last Move\\
8 & Number of Liberties\\
\end{tabular}
\vspace{2mm}

\begin{figure}[h]
\centering
\captionsetup{justification=centering}
\begin{subfigure}{.45\linewidth}
  \centering
  \includegraphics[width=.6\linewidth]{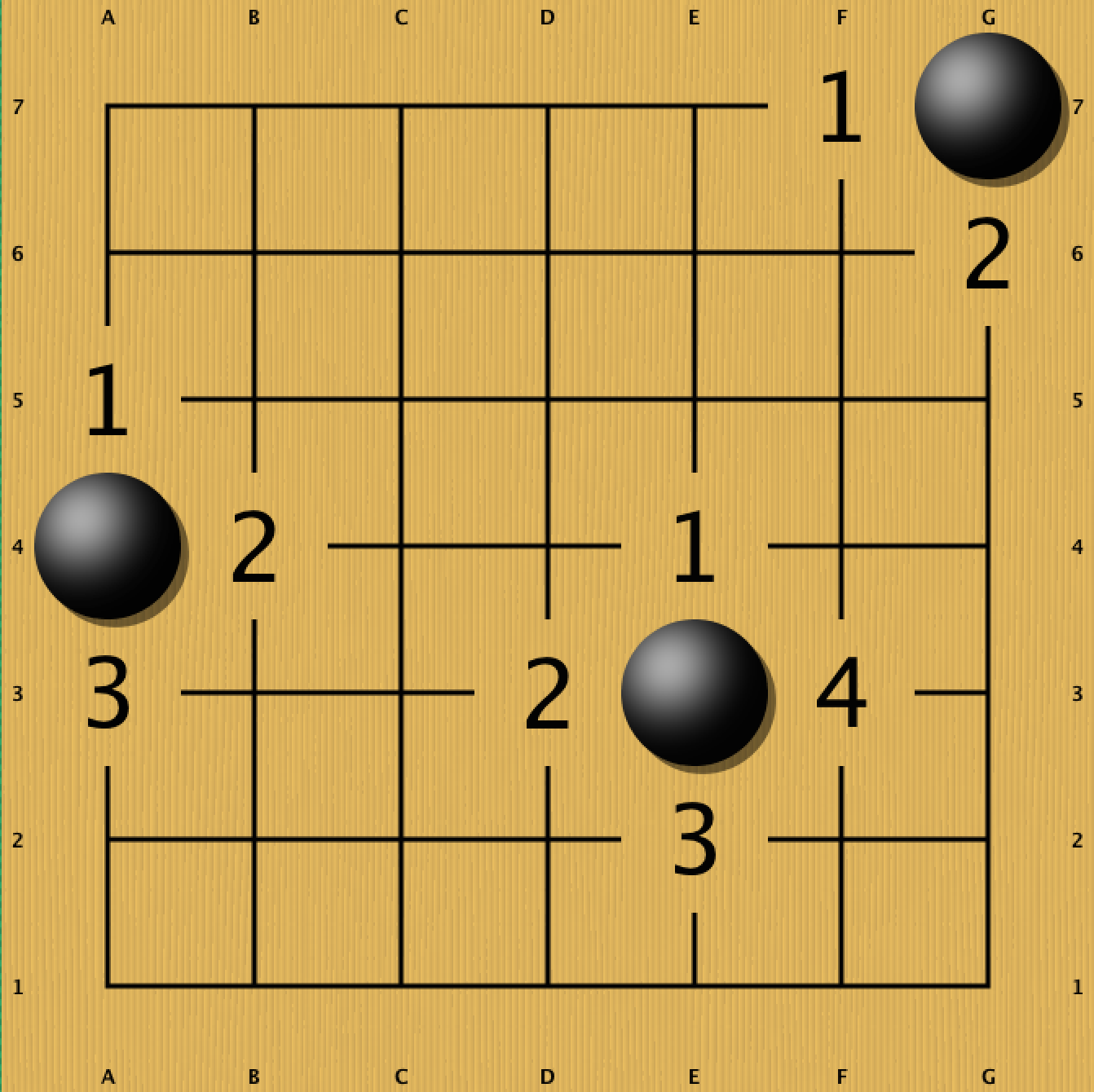}
  \small \caption{Edges do not count as liberties.}
  \label{fig:blsub1}
\end{subfigure}
\ \ 
\begin{subfigure}{.45\linewidth}
  \centering
  \includegraphics[width=.6\linewidth]{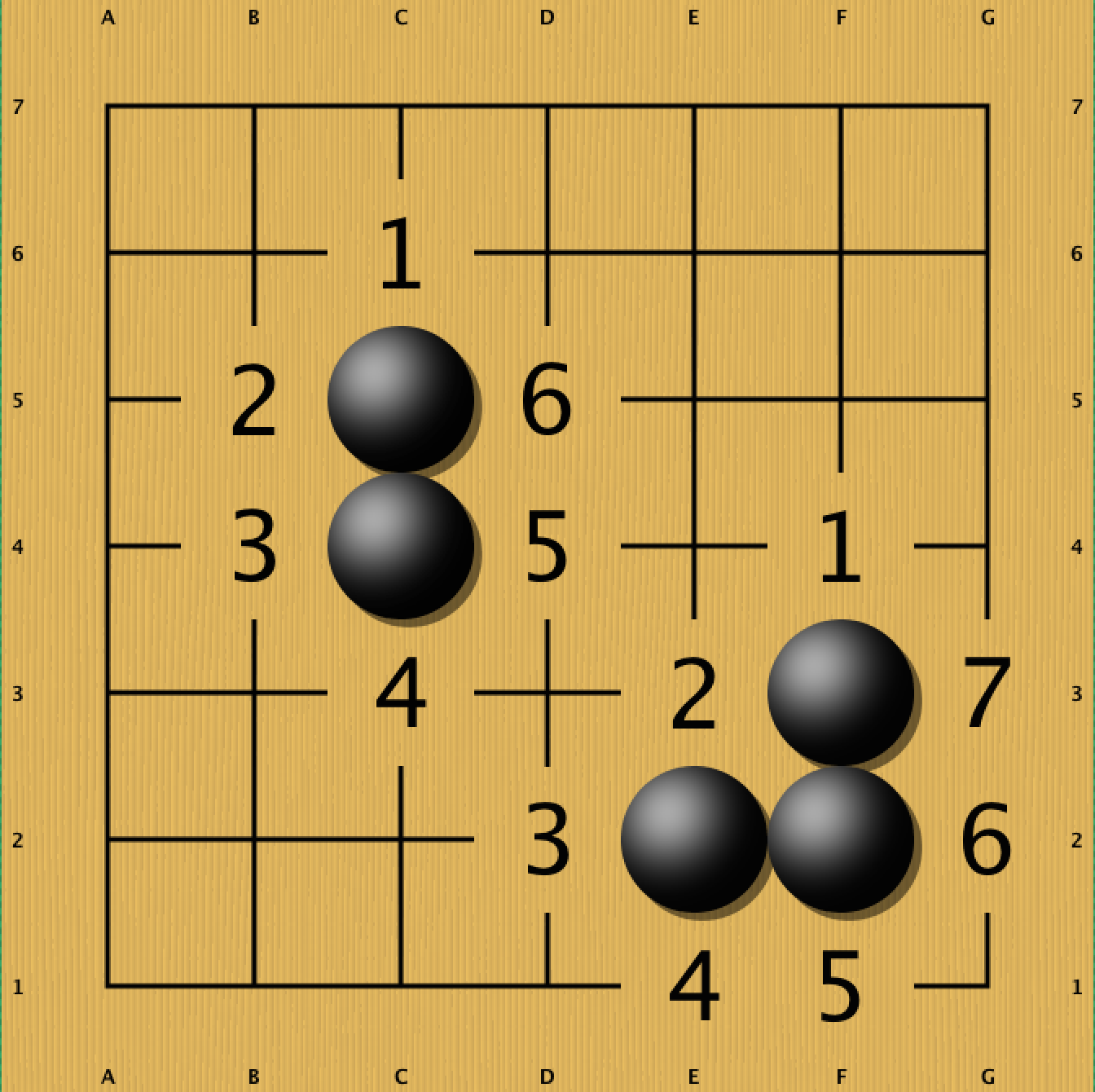}
  \small \caption{Shared liberties within a group only count once.}
  \label{fig:blsub2}
\end{subfigure}
\caption{Number of \textit{liberties} a stone has.}
\label{fig:eglibs}
\end{figure}

The first three layers simply indicate presence of one of our stones, an opponent stone, or a blank space on the $19 \times 19$ board. The fourth layer simply shows the legality of a move to the network.

Features 5-24 follow a slightly different structure. To preserve the non-linearities between different numbers of \textit{liberties} (See Figure \ref{fig:eglibs} for an example of liberties), number of moves since, etc, we opted for a system where a feature would be ``on'' if a stone had one \textit{liberty}, the next feature layer would be ``on'' if a stone had two \textit{liberties}, and so on and so forth. Any stone with more than eight \textit{liberties} would have the eighth feature layer ``on''. A similar idea holds for number of liberties after playing on a certain position, which may well be our only ``lookahead search'', as well as times since move was played, which simply turns on the feature for the point that was played one move ago, two moves ago, etc.

This feature structure is similar to what is used by AlphaGo \cite{SilverHuangEtAl16nature}.

\begin{figure}[h]
\centering
  \includegraphics[width=\linewidth]{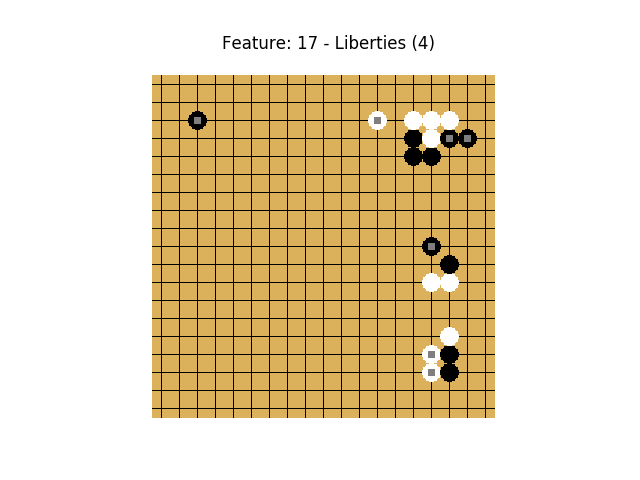}
  \caption{An example board position with all stones with 4 liberties marked. This is the 17\textsuperscript{th} feature layer.}
  \label{fig:featvis}
\end{figure}

We developed a method of visualizing features, as shown in Figure \ref{fig:featvis}. These features, while showing no more than basic rules of the game to the player, such as liberties and legality of moves, allowed the player to look deeper into positions much in the same way that a professional player would.

\section{Results}

Through supervised learning and our novel cross convolutional architecture, we were able to attain a classification accuracy of 47\% on the test set, better than the standard set in \cite{shitversion}. Given the massive amount of test data we had, we were able to achieve this result over 5 epochs (1 million iterations). We also tested the player on a popular go server OGS (Online Go Server), where the player was able to beat 10kyu amateur players. It was also able to beat the well-known bot GnuGo, an MCTS-based algorithm which has a skill level of about 11kyu.

\begin{figure}[h]
\centering
  \includegraphics[width=\linewidth]{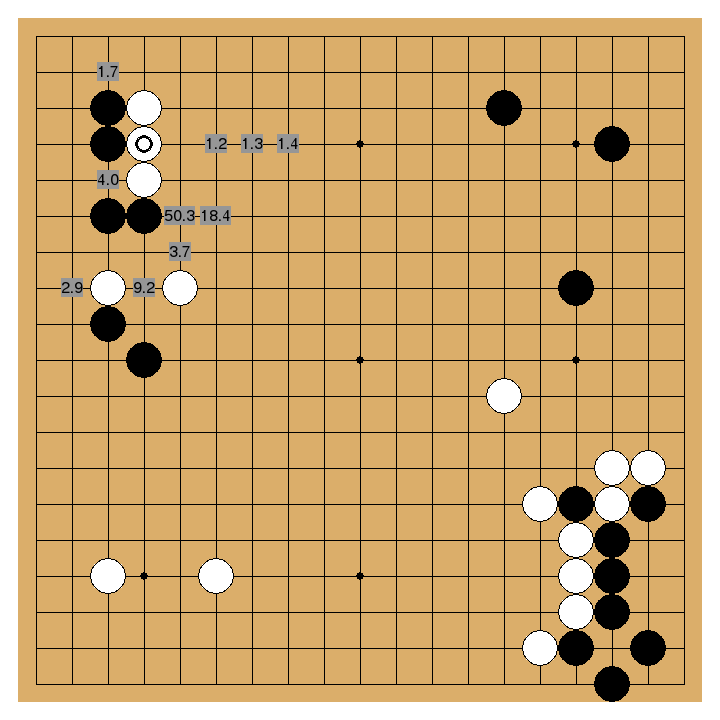}
  \caption{An example position from game 1 of Ke Jie vs AlphaGo, played recently. The move played by Ke Jie (marked with a probability of $50.3\%$), was guessed correctly by our network.}
  \label{fig:model}
\end{figure}

We were even able to evaluate our player by seeing what moves it predicted in the recent Ke Jie vs. AlphaGo matches. Only a couple of the moves played by both players were outside the top 10 considered moves by our network, indicating the strong skill in Go it has acquired by only looking at the current board position.

\section{Conclusion}
We have successfully created a player that is capable of playing intuitively along. Amazingly, we have shown through experimentation that our player is also capable of understanding basic patterns, formations, and playing aggressively. 

As mentioned previously, we were not able to finish implementing reinforcement learning. Therefore, finishing reinforcement learning will be our next milestone to improve our player. Future work also involves expanding the model so it plays more effectively, understands the game at a deeper level, and knows when to pass or resign. 

We also plan to host our player on an online Go server such as OGS or KGS so we can gather more evaluations for our player.

{\small
\bibliographystyle{ieee}
\bibliography{egbib}
}

\end{document}